\documentclass[runningheads]{llncs}

\usepackage[T1]{fontenc}
\usepackage{graphicx}
\usepackage{graphics}
\usepackage{amsmath}
\usepackage{amssymb}
\usepackage{booktabs}
\usepackage{array}
\usepackage{amsmath}
\usepackage{pifont}
\usepackage[x11names]{xcolor}
\usepackage{arydshln}
\usepackage[breaklinks,colorlinks]{hyperref}
\usepackage[table]{xcolor}
\usepackage{colortbl}
\usepackage{comment}
\usepackage[capitalize]{cleveref}
\crefname{section}{Sec.}{Secs.}
\Crefname{section}{Section}{Sections}
\Crefname{table}{Table}{Tables}
\crefname{table}{Tab.}{Tabs.}

\RequirePackage{xspace}
\makeatletter
\DeclareRobustCommand\onedot{\futurelet\@let@token\@onedot}
\def\@onedot{\ifx\@let@token.\else.\null\fi\xspace}

\def\eg{\emph{e.g}\onedot} 
\def\ie{\emph{i.e}\onedot}

\makeatother

\title{Cross-Domain Human Action Recognition from\\ Multiview Motion and Textual Descriptions}

\author{Yannick Porto$^{1,2}$, Renato Martins$^{1}$, \\ Thomas Chalumeau$^{2}$, Cédric Demonceaux$^{1}$}

\authorrunning{Y. Porto et al.}
\titlerunning{Cross-Domain Human Action Recognition with Multiview Motion Cues}

\institute{$^1$Université  Bourgogne Europe, CNRS -- France\\
$^2$TEB Group, Prynel SAS -- France}

\usepackage{fancyhdr}
\fancypagestyle{firstpage}{
  \fancyhf{} 
  \fancyhead[C]{\vspace{-0.5cm}\\ Preprint paper version accepted and presented at\\ 2026 International Conference on Pattern Recognition (ICPR)} 
}

\begin{document}

\maketitle

\begin{abstract}
Robustness to domain changes is a key capability for effective deployment of human action recognition systems in real-world scenarios, where action categories at inference can present important domain shifts or even unseen actions from training. In this context, improving the recognition capabilities of Zero-Shot Action Recognition models (ZSAR), without requiring strong annotation efforts, remains a central challenge. Most ZSAR approaches assume that actions are observed under geometric conditions similar to those seen during training. In practice, variations in human body orientation and camera viewpoint add a significant domain gap in ZSAR, substantially limiting generalization to novel action–motion combinations. In this context, this paper presents a novel orientation-aware action recognition approach with improved cross-domain capabilities. Our approach combines motion cues of multiple camera viewpoints and text descriptions of human actions in the training phase. We present a new orientation-aware motion encoding network to learn different motion features, and adapt a specific orientation-aware text prompt to match the corresponding features at inference. Extensive experiments demonstrate that the proposed method consistently improves ZSAR performance across different recognition benchmarks, outperforming recent state-of-the-art zero-shot approaches on NTU-RGB+D, BABEL, NW-UCLA, and on two surveillance datasets. In addition, the learned representations exhibit strong transfer learning capabilities, yielding competitive performance on both cross-domain and same-domain recognition of seen actions. Code and trained models are available at: \url{https://icb-vision-ai.github.io/OrientationAware-HAR} 
\end{abstract}

\thispagestyle{firstpage}
\section{Introduction}
\label{sec:intro}

\begin{figure}[t]
\centering
\includegraphics[width=0.8\linewidth]{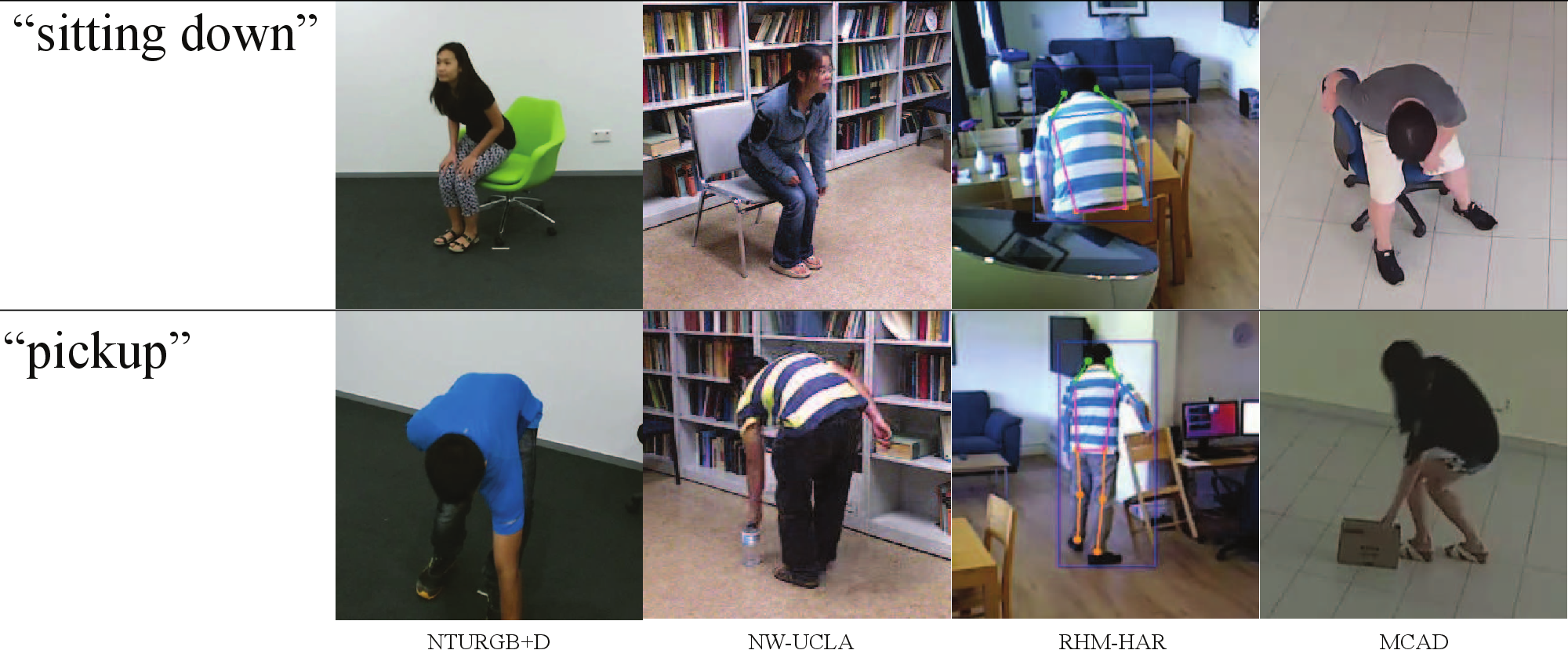}\vspace*{-0.2cm}
\caption{\textbf{Geometric domain gaps actions over four different datasets.} The benchmarks contain sequences with clear distinct camera setups (notably regarding position and viewpoint), as shown for actions ``sitting down'' (top) and ``pickup'' (bottom). These changes are also often observed when deploying models in real contexts.}
\label{fig:datasets}\vspace*{-0.5cm}
\end{figure}

A fundamental challenge in the deployment of Human Action Recognition (HAR) systems, in typical cross-domain real-world conditions, lies in the reliance on large amounts of annotated motion data, which are difficult and costly to obtain for every new deployment domain~\cite{wenhui_li_2017_884592,shahabian2023rh}. In real-world scenarios, action labels are often incomplete, inconsistent, or defined using different vocabularies across datasets and applications~\cite{shahroudy2016NTU,wang2014cross}, making it impractical to anticipate all action categories at training time. For example, an action labeled as ``pick up'' during training may appear at inference in another dataset as ``collect'' or ``gather'', reflecting semantic variations that are rarely aligned across domains. This misalignment between seen and unseen action classes highlights the necessity of Zero-Shot Action Recognition (ZSAR) methods, which aims to recognize actions without requiring additional labeled motion samples. Beyond semantic discrepancies, ZSAR in cross-domain settings is further complicated by geometric domain shifts in motion observations: training data are often captured from canonical or frontal viewpoints, while test-time sequences may be observed from side or back views (as depicted in~\cref{fig:datasets}). As a result, even seen action classes at training are affected by geometric domain shifts, leading to compounded challenges that are frequently not adequately addressed by conventional classification-based HAR methods~\cite{Yan2018SpatialTG,Duan2022} or existing ZSAR methods~\cite{mehraban2024stars,yu2024exploring,chen2025neuron}. In this context, aligning motion representations with textual descriptions emerges as a natural and powerful mechanism to bridge semantic gaps between action labels across domains. Recent works have demonstrated the effectiveness of text–motion alignment~\cite{tevet2022motionclip,yu2024exploring} for human action understanding, learning joint embedding spaces that capture fine-grained correspondences between motion patterns and language semantics. However, these approaches are still affected by geometric domain shifts even under same-domain assumptions~\cite{li2024sadvae,zhu2024part}, where the set of action classes remains the same between training and testing.

To address these challenges, we introduce a multi-view training strategy that learns representations of human actions from multiple, independently observed body orientations, paired with matched orientation-aware textual descriptions. As a prerequisite to this design, we distinguish between two broad categories of HAR data sources: (1) systems operating under controlled conditions and a large number of calibrated sensors (such as of existing motion capture systems), and (2) data derived from pose estimation in videos captured \textit{in the wild} using single(-few) and uncalibrated sensors. This distinction reveals a substantial geometric discrepancy between training and deployment conditions. Motivated by the observation of viewpoint-dependent performance of existing systems, we propose a novel orientation-aware multi-view approach that bridges the geometric gap between these two paradigms. Traditionally, the multi-view problem in action recognition has been addressed through the \emph{consensus principle}, with particularly early-fusion frameworks involving 3D pose estimation, depending on complete and accurate visibility of body parts across views. It involves limitations, arising primarily from the dependence on precise inter-camera calibration and temporal tracking of body keypoint motions. In contrast, we advocate for a \emph{complementary principle}, wherein diverse and potentially incomplete viewpoints contribute by distinct, non-overlapping information that, when aggregated, enhance the robustness and discriminative power of the action recognition system. Our method integrates independently learned view-specific representations at a higher level, mitigating the impact of partial occlusions and action ambiguity. This ensemble of perspectives enables feature orientation awareness and the compensation for missing or corrupted data in one view with complementary cues from other views.
This design not only improves the model's transfer learning capabilities across domains, but also in same domain contexts showing the effectiveness of multi-view systems in complex, real-world scenarios. We evaluate the method in five different benchmarks to demonstrate the applicability of the model on real-world sequences for both same and cross-domain contexts, including two datasets on surveillance and monitoring scenarios. This work brings three main contributions:
\begin{itemize}
    \item We design a body orientation-aware action classification network, composed of two complementary attention branches, that leverages body orientation cues in a positional encoding for guiding the learning of motion features with viewpoint information. The proposed strategy outperforms state-of-the-art recognition models on different benchmarks even when only a single viewpoint is available at inference.
    \item In order to further improve performance when dealing with different domains, we present a joint classification with the alignment of features from motion and textual description of actions enriched with viewpoint information. In this context, text action prompts are enriched with viewpoint information. The proposed method shows improved classification results in cross-domain experiments in five widely adopted benchmarks, even overpassing the current state-of-the-art zero-shot method in all considered benchmarks (NTU-RGB+D, NW-UCLA, RHM-HAR and MCAD).
    \item It is important to highlight that our method requires multiview and text cues solely in the training stage, which is done with renderings of virtual camera viewpoints and from existing textual actions labels. During inference, the method can be also used with a single camera viewpoint as shown in all experiments. 
\end{itemize}

\section{Related Work}\label{sec:rel}

\paragraph{Multi-View Action Recognition.}
Research on multi-view human action recognition using skeleton data remains limited. Existing strategies predominantly focus on achieving viewpoint invariance ~\cite{zhang2019view,Sun2020,Zhao_2021_CVPR,hou2022shifting,wang2022temporal,yang2024view}. Zhang et al.~\cite{zhang2019view} present a dual-network framework comprising a view-adaptive RNN and CNN. Each network learns to optimally reorient the 3D skeleton sequence, and their outputs are fused for improved recognition. However, this method does not explicitly exploit inter-joint relationships inherent to the skeleton structure. Hou et al.~\cite{hou2022shifting} address this limitation by proposing a multi-head architecture that processes multiple orientations of input skeletons. A skeleton anchor proposal module generates spatial anchor points and computes angular metrics relative to joints, thereby enhancing structural awareness across viewpoints. We take inspiration from their data augmentation technique for the design of ours but we take into account the body self-occlusions with a projection component. Wang et al.~\cite{wang2022temporal} approach viewpoint variation via a novel extension of Dynamic Time Warping. Their method jointly aligns temporal sequences and simulated camera viewpoints by modeling smooth paths between query and support frames. Other approaches pursue viewpoint invariance through data augmentation. Yang et al.~\cite{yang2024view} train a visual encoder on a motion retargeting objective to fuse viewpoint and character attributes from different sequences. The generated hybrid data subsequently facilitates transfer learning for action recognition. Additionally, contrastive learning has been adapted to improve robustness under occlusion~\cite{yang2021Occlusion} or forcing the features coming from different views to be closed to each others   \cite{Shah2023MultiViewAR,Siddiqui_Tirupattur_Shah_2024}. {These mentioned multi-view methods show promising results with data augmentation techniques for view-invariant action recognition but they force the network to follow a consensus principle and do not exploit the capabilities of multiple views during inference, where each viewpoint brings a different information. On the contrary, we condition the weights of the network on this viewpoint position. They also require fine-tuning with a large set of training examples in the targeted domain and do not demonstrate a domain change ability.}

\vspace{-0.5cm}

\paragraph{Motion-Text Combination for Action Recognition.}
\label{subseq:rel-cm}
Recent approaches to action recognition have increasingly leveraged cross-modal information such as visual, motion and text data. Matching a motion with a specific action is then not dependent on the set of actions solely defined during training. Instead, textual descriptions of movements are projected into a shared embedding space aligned with motion features. For example, MotionCLIP~\cite{tevet2022motionclip} aligns the latent spaces of text, motion, and images using a transformer trained with both feature similarity and pose reconstruction objectives. Although these models are evaluated on standard action recognition benchmarks, several works~\cite{li2024sadvae,mehraban2024stars,zhu2024part} also report performance in Zero-Shot Learning (ZSL) and generalized ZSL settings, where models are tested on unseen or in a mix of seen and unseen classes, respectively, while training is done to a subset of the classes. Most of these methods~\cite{zhu2024part,mehraban2024stars,yu2024exploring,chen2025neuron} exploit Large Language Models (LLMs) to generate enriched class descriptions, thereby enhancing the performance of the text encoder. We take inspiration from these works for our cross-domain portability, however we extend the cross-modal alignment to consider multi-view orientation cues in the action description. Our approach is designed to maintain performance in \emph{in-the-wild} action recognition contexts typically present in surveillance and monitoring contexts, as evaluated in the experiments in two surveillance datasets.

\section{Method}
\label{sec:method-training}

This section presents our action recognition method taking into account multi-view information in the training and available textual action descriptions extended with body orientation cues. Our method comprises four main components, as shown in the schematic representation in~\cref{fig:training}. A major insight is to condition the feature extraction of the motion (``Orientation-Aware Network'') with the actor orientation, considering different virtual camera viewpoints (``Projection Component''). The projection is done solely in the training phase. Then, available action textual descriptions (``Action Description'') are leveraged to guide the feature learning and align the motion and textual feature maps for cross-domain capability. The proposed model effectively handles features with further robustness to viewpoint changes, and is also adapted to cross-domain ZSL applications with unseen actions.

\subsection{Motion Projection with Virtual Viewpoints}
\label{subsec:projection}

\begin{figure}[t]
    \centering
    \includegraphics[width=\linewidth]{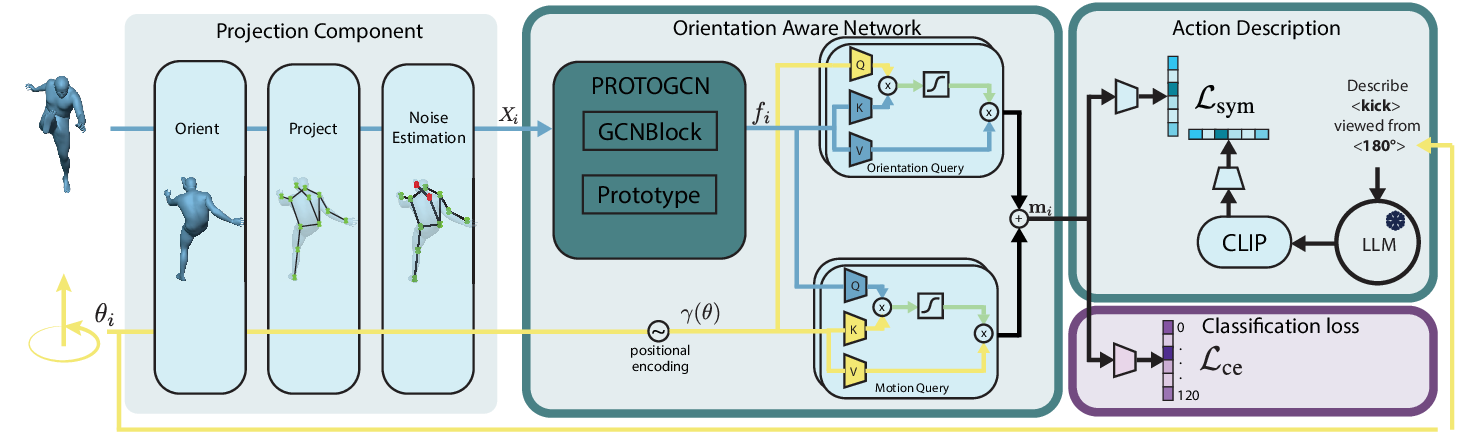}
    \caption{\textbf{Overview of our action recognition training pipeline.} First, the ``{Projection Component}'' generates virtual rendered views by projecting the motion sequence in virtual camera viewpoints. These projected motions are then passed to the ``{Orientation Aware Network}'' which encodes the motion and condition the extracted features with the given body orientation angle thanks to a dual-branch attention mechanism. The ``{Action Description}'' generates text descriptions from prompts containing the corresponding action label and body orientation information. Finally, the network is trained with a joint contrastive and classification losses to guide the learning of the final features with the alignment of action descriptions and multi-view motion features.} \vspace*{-0.3cm}
    \label{fig:training}
\end{figure}

For the training, we consider as input 3D motion sequences with known SMPL~\cite{SMPL} parameters. The first stage of our training scheme is then to obtain projected 2D skeleton sequences, by simulating virtual cameras with known parameters (as illustrated in the left side of \cref{fig:training}). This allows obtaining a diverse set of 2D motions with different joint visibility from a single motion sequence. As real-world sequences processed at evaluation might contain self-occlusions, following recommendations in \cite{Duan2022}, we simulate this effect by checking if the rays passing from the camera position to the limb joints of the skeleton cross the plane composed of the torso joints (see left of \cref{fig:training}). This addition of occlusion and multi-view variability aims to bring the motions at training closer to \textit{in the wild} conditions.

To generate the virtual camera rig of viewpoints, we split the range of yaw angle in twelve directions as in CCTV cameras. We then associate a viewing angle with the closest orientation of the hips of the 3D skeleton.
The reference viewpoint corresponds to the virtual camera viewpoint with an optical axis orthogonal to the hips vector in the first frame of the sequence. This projection module provides two inputs to our network: a sequence of joints $X_i$ projected on the $i_{th}$ view, as well as a unique body orientation $\theta_i$.

\subsection{Orientation-Aware Network}

Our orientation-aware (OA) motion extraction network is composed of two branches based on multi-head cross attentions. First, the body orientation $\theta \in [-\pi,\pi]$ is mapped into a continuous representation $(\sin(\theta), \cos(\theta))$. Inspired by~\cite{mildenhall2021nerf}, for conditioning the orientation information in a higher dimensional space, we represent this orientation with a positional encoding at different frequency levels:
\begin{align}
\label{eq:pe}
    \gamma(\theta) = ( \sin(2^0 \pi \theta), \cos(2^0 \pi \theta), \cdots, \\ \notag \sin(2^{L-1} \pi \theta), \cos(2^{L-1} \pi \theta) ).
\end{align}
We have selected $L=192$ as being half of motion feature dimension. This orientation is used to condition the learning of the spatio-temporal encoder $E$ of the motion which is used to extract $f_i \in \mathbb{R}^{C_i}$ with $C_i$ features, out of each projected skeleton sequence into \( i \)-th view as $f_i = E(X_i)$. We chose ProtoGCN~\cite{liu2025revealing} as spatio-temporal encoder for its ability to handle joints relation. Its selection followed a comparison with alternative encoders regarding the specific contribution of orientation features.
The orientation condition of the network is then done with two complementary branches (as depicted in \cref{fig:training}):

\begin{itemize}
    \item \emph{Orientation as query}: The upper branch of the network is responsible to condition the pose sequence based on the perceived body orientation to answer the question: ``From this angle, what can we say from the motion?''. For this, the normalized orientation angle is projected to the query of the first multi-head attention with an MLP that we couple to a learnable query vector. 
    The key and value are made of the pooled and projected motion features from the ProtoGCN encoder backbone.
\item \emph{Motion as query}: On the second branch (lower multi-head attention), the query is the projected motion features, while the key and value are the projected orientation angle. The resulting motion features responds to ``What can we say about this viewpoint angle from this motion?''. 
\end{itemize}
Both features are then concatenated to represent the final motion features $\mathbf{m}_i$ from the orientation-aware component.
These motion features are conditioned on body orientation and go in the direction of the  ``Complementary Principle'', by getting different information from each view.

\subsection{Action Description Features}
\label{method:description}

At inference time, the subject may perform actions with varying characteristics from the training set (\eg, different motion style, speed or body orientation) and even perform actions not considered in the training. To handle such conditions, we design a cross-modal alignment to match different labels describing a same movement (such as ``phone" or ``telephone call'') as well as unseen possible actions in new domains. The spatial-awareness is provided in the description by considering the orientation of the observed motion in the action textual prompt. For example, if the person is clapping from a back orientation, we want the description to focus more on the elbows movement than the hands, which are certainly occluded.

To leverage datasets with single-word class labels, we enhance these labels using GPT-3.5 Large Language Model (LLM). Our prompting strategy follows prompting schemes presented in~\cite{wei2022chain}, and provides contextual guidance to generate detailed action descriptions. Given that our action recognition framework incorporates viewpoint information, the prompts explicitly specify the orientation of the subject relative to the camera, allowing the model to account for occlusions, viewpoint variability and emphasize into the most visible body parts. The LLM is instructed to produce action descriptions following the template:
“A person $<$action$>$, $<$a brief description of the principal limbs involved from the specified viewpoint$>$, $<$a short description of the subject’s orientation with respect to the camera$>$.”
This process yields a lookup table containing viewpoint-specific action descriptions. The right side of \cref{fig:training} illustrates the proposed cross-modal training pipeline. For text to visual encoding, we employ CLIP~\cite{radford2021learning}, which has demonstrated strong performance in aligning textual and visual modalities. CLIP also facilitates training on long-tailed distributions, as rare class labels can support effective alignment with other modality pairs. Both textual and motion representations are projected into a common embedding space with a fully connected layer.

\subsection{Contrastive Motion-Text Feature Alignment}

To learn features combining motion and corresponding textual descriptions, we adopt a contrastive learning strategy~\cite{balntas2016learning} with pairs of positive and negative motion-text samples. The objective is to learn aligned motion and action features from the similarity between matching pairs, while pushing apart non-matching motion-text pairs. 
Given a positive motion-text pair $(\mathbf{m}_i^+, \mathbf{t}_i)$ and a corresponding negative pair $(\mathbf{m}_i^-, \mathbf{t}_i)$, we first normalize all embeddings to lie on the unit hypersphere: $
\hat{\mathbf{m}} = \frac{\mathbf{m}}{\|\mathbf{m}\|}$, and $\hat{\mathbf{t}} = \frac{\mathbf{t}}{\|\mathbf{t}\|}$, to compute the similarity distance $d(\hat{\mathbf{m}}, \hat{\mathbf{t}}) \in [0, 1]$:
$
d(\hat{\mathbf{m}}, \hat{\mathbf{t}}) = 1 - {(1 + \cos(\hat{\mathbf{m}}, \hat{\mathbf{t}}))}/{2}.
$
The contrastive loss function for a batch of $N$ positive and negative pairs is then defined as:
\begin{align}\label{eq:sym}
\mathcal{L}_{\text{sym}} = \frac{1}{2N} \sum_{i=1}^{N} y_i  d(\hat{\mathbf{m}}_i, \hat{\mathbf{t}}_i)^2 ~ + ~~~~~~~~~~~~~~~~~~~~~~~~~  \\ \notag  + ~ (1 - y_i)\left[ \max\left(0, \mu - d(\hat{\mathbf{m}}_i, \hat{\mathbf{t}}_i) \right) \right]^2,
\end{align}
where $y_i = 1$ if $(\hat{\mathbf{m}}_i, \hat{\mathbf{t}}_i)$ is a positive pair, and $y_i = 0$ otherwise. The margin hyperparameter $\mu$ controls the minimum allowable distance between non-matching motion-text pairs. Each batch is constructed by concatenating $N$ positive motion-text pairs and $N$ negative pairs $\mathbf{M} = [\hat{\mathbf{m}}_1^+; \hat{\mathbf{m}}_1^-; \dots; \hat{\mathbf{m}}_N^+; \hat{\mathbf{m}}_N^-]$ and $\mathbf{T} = [\hat{\mathbf{t}}_1; \hat{\mathbf{t}}_1; \dots; \hat{\mathbf{t}}_N; \hat{\mathbf{t}}_N]$, which have the corresponding labels $\mathbf{y} = [1, 0, \ldots, 1, 0]^\top$.

\paragraph{Training loss.}
During training, we also include a motion classification loss to guide the feature learning for same domain classification, as shown in the bottom right corner of \cref{fig:training}. The motion features are fed into a two layer MLP classifier, and the obtained class probabilities are fed to a standard cross-entropy loss $\mathcal{L}_{\text{ce}}$ term. The full training loss is therefore the weighted combination of the contrastive term (\cref{eq:sym} and motion classification as:
\begin{align}
\label{eq:loss}
\mathcal{L} = \lambda\mathcal{L}_{\text{ce}} + (1-\lambda)\mathcal{L}_{\text{sym}},
\end{align}
\noindent where the hyperparameter $\lambda$ adjusts the contribution of the two terms.

\subsection{Single View and Multiview Inference}

As described in \cref{sec:rel}, most previous works exploit the multiple view information at training time to get an encoder invariant to viewpoint changes, but disregard this information in inference time even when available as in CCTV. Our model, on the other side, can also benefit from viewpoints available at inference time to improve the action recognition. In the experiments, we provide the results of our model in both single-view (SV) and multi-view (MV) inference settings. MV is computed with the mean of each orientation probabilities, while for SV we choose only one orientation for the inference.
We highlight that at inference time, the proposed method can be used with multiple views, \eg, with a rig of cameras, but also in single view settings, \ie., with a single camera. When multiple viewpoints are available at inference, the final classification of action is the average of the probabilities for each view. \\

\vspace{-0.3cm}
\noindent\emph{Cross-domain inference.} For the cross-domain experiments with text descriptions, we compute the cosine similarity scores between the motion feature for view $i$ and all possible action description features as:
\begin{align}
    \mathbf{s}_i = \hat{\mathbf{m}}_i \cdot \hat{\mathbf{T}}^\top \in \mathbb{R}^K,
\end{align}
with $K$ the number of possible action classes.
The final output is then the average of scores across all $N$ views: $
    \bar{\mathbf{s}} = \frac{1}{N} \sum_{i=1}^N \mathbf{s}_i,$
and the action class $\hat{k}$ is simply then $\hat{k} = \arg\max_k \bar{s}[k]$. 

\vspace{2cm}

\section{Experiments}
This section starts with information on the adopted datasets, competitors and experimental setup. We then discuss the obtained action classification results.

\paragraph{Baselines.} We have selected several zero-shot approaches for cross-domain evaluation: CADA-VAE~\cite{schonfeld2019generalized}, JPoSE~\cite{wray2019fine}, ReViSE~\cite{hubert2017learning}, DeViSE~\cite{frome2013devise}, PURLS~\cite{zhu2024part}, the state-of-the-art approaches Neuron~\cite{chen2025neuron} in zero-shot learning, and ViA~\cite{yang2024view} in multiview transfer learning. For comparison with ZSL methods, ViA was coupled with a CLIP text encoder in zero-shot experiments.
We also considered recent competitors leveraging motion and textual descriptions for action recognition with single-transformer-based backbones~\cite{petrovich2023tmr,tevet2022motionclip}, as well as MotionPatches~\cite{yu2024exploring} with ViT-based backbone.
For the ablation and sensitivity experiments, we have selected two different spatio-temporal encoders for comparison with the original results in our experiments: i) 2s-AGCN \cite{Shi2sAGCN} which comes with pretrained weights and results for different benchmarks and ii) ProtoGCN \cite{liu2025revealing} which is a recent state of the art action classification.

\paragraph{Datasets.} We have selected three widely adopted human action classification benchmarks NTU-RGB+D~\cite{shahroudy2016NTU,liu2020ntu}, NW-UCLA~\cite{wang2014cross} and BABEL~\cite{BABEL:CVPR:2021} for the evaluation, and two recent datasets for the real applications: RHM-HAR~\cite{shahabian2023rh} which contains robot and surveillance views and MCAD~\cite{wenhui_li_2017_884592} recorded from a real multiview CCTV system. Some examples are presented in \cref{fig:datasets}, where the multiple views of the different domains record a same action from different performer angles. In the presented experiments, we have adopted the same setups described in \cite{zhu2024part}: 30/30 classes split for NTU-RGB+D-60, where 30 classes are seen and 30 other classes are unseen at training; and 5/5 split of seen and unseen classes for NW-UCLA. For MCAD, we use the provided cross-subject, cross-view and 9/9 cross-action splits. Finally, as a downstream monitoring application, we have considered {RHM-HAR} which contains 4 synchronized viewpoints with 3 opposed surveillance cameras and 1 camera embedded on a moving robot. We create a 7/7 actions split and a views split with the robot view as training samples and the CCTV views as test samples to evaluate our system on this application.

\begin{table}[t]
    \vspace{0pt}
    \begin{minipage}{0.45\textwidth}
        \vspace{0pt}
        \caption{\textbf{Cross-domain top1 evaluation with ZSL.} Done on NTU-RGB+D-60 and NW-UCLA.} 
        \vspace{0.2cm}
        \label{tab:cross-domain}
        \resizebox{\linewidth}{!}{
            \begin{tabular}{lcc}
            \toprule
            Method & NTU-60 & NW-UCLA\\
            \midrule
            ZSL Setup & 30/30 & 5/5\\
            \hdashline
            CADA-VAE~\cite{schonfeld2019generalized} (CVPR'19) & 11.51 & --\\
            JPoSE~\cite{wray2019fine} (ICCV'19) & 14.81 & --\\
            ReViSE~\cite{hubert2017learning} (ICCV'17) & 12.39 & 44.99\\
            DeViSE~\cite{frome2013devise} (NeurIPS'13) & 18.48 & 36.02\\
            PURLS~\cite{zhu2024part} (CVPR'24) & 23.52 & 49.47\\
            Neuron~\cite{chen2025neuron} (CVPR'25) & 24.84 & --\\
            ViA~\cite{yang2024view} (IJCV'24) & \underline{25.02} & \underline{54.47}\\
            Ours & \textbf{28.48} & \textbf{69.09}\\
            \bottomrule
            \end{tabular}
        }
    \end{minipage}
    \hfill
    \begin{minipage}{0.45\textwidth}
        \vspace{-0.9cm}
        \caption{\textbf{Cross-domain top1 evaluation with ZSL and ZSCD.} Done on four different datasets.} 
        \vspace{0.2cm}
        \label{tab:cross-domain-spaced}
        \resizebox{\linewidth}{!}{
            \begin{tabular}{lcccc|c}
            \toprule
            Method & NTU-60 & NW-UCLA & RHM & MCAD & mean\\
            \midrule
            ZSL & 30/30 & 5/5 & 7/7 & 9/9 &\\
            \hdashline
            ViA~\cite{yang2024view} & 25.02 & 54.47 & 29.62 & 18.70 & 31.95\\
            Ours & \textbf{28.48} & \textbf{69.09} & \textbf{52.03} & \textbf{50.12} & \textbf{49.93}\\
            \midrule
            ZSCD & 0/30 & 0/5 & 0/7 & 0/9 &\\
            \hdashline
            ViA~\cite{yang2024view} & 9.06 & 38.72 & 17.38 & 14.16 & 19.83\\
            Ours & \textbf{26.17} & \textbf{56.12} & \textbf{41.67} & \textbf{31.29} & \textbf{38.81}\\
            \bottomrule
            \end{tabular}
        }
    \end{minipage}
\end{table}

\vspace{1cm}

\paragraph{Experimental setup.}
We trained the proposed method described in \cref{sec:method-training} with motion samples from BABEL. For all experiments the adopted margin in the constrastive loss is $\mu = 0.5$, $\lambda = 0.5$ for pretraining the weights, and $\lambda = 1$ in finetuning. GPT3.5 is used as LLM to generate the action descriptions.
The preprocessing steps at training are as follows:
\begin{itemize}
    \item Random uniform sampling~\cite{Duan2022} is applied to the provided sequence to obtain sequences with length of 150 frames.
    \item The twelve virtual views are generated with the process detailed in \cref{subsec:projection}, which corresponds to body orientations from $-180$° to $150$° with a uniform step of $30$°.
    \item Following~\cite{chang2019poselifter}, additive noise with a mixture model consisting of Gaussian and uniform distributions is used to produce corrupted 2D skeleton sequences.
\end{itemize}
The model optimization was done with Adam using a learning rate of $0,001$, learning rate decay of $0,1$ at epoch $30$, and batch size of $16$. The training takes around 12 hours in a GPU Nvidia A5000Ada.

\subsection{Cross-Domain Action Classification} \label{exp:cd}

A core feature of action recognition algorithms is to be robust and employable on real-world sequences. After training, the multiview projection component is not kept (left side of \cref{fig:training}).
For the pose estimation at inference, we have adopted~\cite{cheng2020bottom} in order to be consistent with the COCO skeleton convention used in training time. 
For the human body orientation estimation (HBOE) used for the orientation conditioning, we have adopted the same backbone from~\cite{aaai_ZhaoLYLZRCW24}. The orientation is estimated taking into account cropped images of the persons as input.
At inference time, we want to find the action performed by the subject out of a list of actions. Some of these actions can match the list during training, but some can be new unseen actions. Therefore, this new list is passed into the language processing pipeline which gives features to be compared to the motion ones. Text features with closest similarity indicate the performed action.\\

\begin{table*}[t]
    \caption{\textbf{Multiview pretraining influence.} Comparison of different pretrained networks on cross-subject (CS) and cross-view (CV) splits of four datasets. \textbf{Bold} and \underline{underlined} text are for first and second best top1 respectively.}
        \label{tab:transfer-learning}
        \resizebox{\linewidth}{!}{
        \centering
            \begin{tabular}{llccccccc}
            \toprule
            Pretraining Data & Settings & Method & \multicolumn{2}{c}{NTURGB+D-60} & NW-UCLA & RHM-HAR & \multicolumn{2}{c}{MCAD} \\
            \midrule
            & & & CS & CV & CV & CV & CS & CV\\
            \midrule
            Posetics (142k samples) & SV & ViA~\cite{yang2024view} & 89.6 & 96.40 & \underline{82.72} & 76.86 & 93.17 & 86.93\\
            Posetics (142k samples) & SV & Ours & \underline{90.91} & \underline{97.05} & 80.34 & \underline{82.06} & \underline{93.52} & \underline{90.94} \\
            Babel (44k samples) & SV & Ours & 90.74 & 97.03 & 79.48 & 76.72 & 94.19 & 90.59\\
            Babel (44k samples) & MV & Ours & \textbf{93.77} & \textbf{97.54} & \textbf{83.59} & \textbf{83.79} & \textbf{94.55} & \textbf{91.17}\\
            \bottomrule
            \end{tabular}\vspace*{-0.3cm}
        }
\end{table*}

\vspace{-0.3cm}
\noindent\emph{Results.} 
The cross domain experiments are shown in~\cref{tab:cross-domain} and \cref{tab:cross-domain-spaced} considering BABEL-60 for training, and test on NTU, NW-UCLA, RHM-HAR and MCAD benchmarks. We perform two different experiments: one is a Zero-Shot Learning (ZSL) (as described in \cref{subseq:rel-cm}), where $n_s$ classes are seen at training in the same domain than the $n_u$ unseen classes at inference. The second experiment is a Zero-Shot Cross-Domain (ZSCD) where no finetuning is performed on the targeted domain, and we only keep the pretrained weights from the source domain to infer the $n_u$ classes.

\cref{tab:cross-domain} highlights the ability of our approach to overpass other zero-shot methods on two widely adopted datasets. A gain of $+3.46\%$ and $+14.62\%$ are provided to the NTURGB+D and NW-UCLA splits compared to the pretrained network ViA~\cite{yang2024view}, which is the state-of-art method presenting the second best performance in the evaluation. We report their results with ours on other benchmarks and ZSCD in \cref{tab:cross-domain-spaced}. Although ViA have knowledge about half of the targeted domain in ZSL experiment, their results still remain under the accuracy of our model in ZSCD (accuracy of $26.1\%$ on 30 classes in NTU and $56.1\%$ on 5 classes of NW-UCLA). Finetuning the model on the $n_s$ remaining classes allows our network to even reach $28.5\%$ and $69.1\%$ top1 accuracy on both domains. The same experiments are operated on RHM-HAR and MCAD datasets which contain high camera displacements, with back and side views, compared to other laboratory data with front-faced subject sequences. Please notice, the method performance is twice more accurate than ViA in ZSCD. This suggests that our network and training pipeline allow a more effective transfer to other domains represented by these different datasets.

\begin{table}[t]
    \caption{\textbf{Same domain evaluation.} Action classification comparisons on BABEL-60 with state-of-the-art approaches. Compared to known results on BABEL-60 our network and training pipeline allow a gain of $+5.6\%$ on top1 and $+7.2\%$ on top5.}
       \label{tab:soa}
        \centering
            \begin{tabular}{lcccc}
            \toprule
            Method & Data & Features & \multicolumn{2}{c}{BABEL-60}\\
            & & & top1 & top5\\
            \midrule
            TMR \cite{petrovich2023tmr}& 3D & motion+text & 30.13 & 41.52 \\
            MotionClip \cite{tevet2022motionclip} & 3D & motion+text & 40.9 & 57.71 \\
            2s-AGCN~\cite{Shi2sAGCN} & 3D & motion & 41.14 & 73.18 \\
            MotionPatches \cite{yu2024exploring} & 2D & motion+text & 41.33 & 68.97 \\
            ViA \cite{yang2024view} & 2D & motion & 41.79 & 71.37 \\
            \rowcolor{green!20}
            \midrule
            Ours & 2D  & motion+text & \textbf{47.41} & \textbf{80.40}\\
            \bottomrule
            \end{tabular}
            \vspace{0.4cm}
    \caption{\textbf{Ablation study of the multiview projection and orientation aware components.} This ablation analyses the contribution from each of the motion designings, with two different backbones. The different modules bring gains of accuracy alone and combined to these motion encoders.}\vspace{0.2cm}
        \label{tab:classification}
        \resizebox{0.6\linewidth}{!}{
        \centering
            \begin{tabular}{lccc|cc}
            \toprule
            Method & Backbone & Projection & Orientation & \multicolumn{2}{c}{BABEL-60}\\
            \midrule
            & & & & top1 & top5 \\
            \midrule
            \multicolumn{4}{c|}{\textbf{Single-view inference}} & & \\
            \hdashline
            2s-AGCN~\cite{Shi2sAGCN} & 2s-AGCN & & & 41.14 & 73.18 \\
            & 2s-AGCN & \checkmark & & 45.75 & 77.80 \\
            & 2s-AGCN & \checkmark & \checkmark & 46.06 & 78.29\\
            & Protogcn & \checkmark & & 45.88 & 78.12 \\
            Ours & Protogcn & \checkmark & \checkmark & 46.13 & 78.35\\
            \midrule
            \multicolumn{4}{c|}{\textbf{Multi-view inference}} & & \\
            \hdashline
            & 2s-AGCN & \checkmark & & 46.54 & 79.12 \\
            & 2s-AGCN & \checkmark & \checkmark & 47.19 & 79.48\\
            & Protogcn & \checkmark & & 47.03 & 79.39 \\
            Ours & Protogcn & \checkmark & \checkmark & 47.41 & 80.40 \\
            \bottomrule
            \end{tabular}
        }
\end{table}

\subsection{Multi-view Action Classification} \label{exp_cm}

The contribution of our multiview pretraining pipeline is compared with the state-of-the-art pretraining method ViA~\cite{yang2024view} in ~\cref{tab:transfer-learning}. The results show that even large datasets employed at pretraining, such as Posetics~\cite{yang2021unik} which is at least three times larger than BABEL, do not overpass the accuracy obtained with our multiview pipeline. We provide results obtained with different data and settings: Posetics and BABEL in SV and MV settings which are samples with and without multiview projections and loss (Posetics MV being not considered because of the single view recorded nature of the data). This shows that our network can also get a higher accuracy with the same data than competitors. Secondly, our MV setting allows to overpass their results with fewer samples in the pretraining.

\paragraph{Same domain.}
We also compare the models for performing classification in the same domain. These experiments are notably done to evaluate the contribution of our multiview training and inference strategy regarding other existing approaches when the source dataset is the same than the targeted one without pretraining the model. Our overall results of human action classification on BABEL are highlighted in \cref{tab:soa}. These results show that our approach overpasses by a large margin several of recent competitors in the top1 and top5 metrics on the BABEL-60 dataset. These results include the projection, the orientation-aware network, as well as the language component combined with the classification output. We have obtained a gain of $+5.6\%$ on top1 and $+7.2\%$ on top5 (\ie, a relative improvement of 13\% and 10\% respectively).

\begin{figure}[t]
\begin{minipage}[t]{0.2\linewidth}
\vspace{0pt}
\centering
\fbox{\includegraphics[height=1\linewidth]{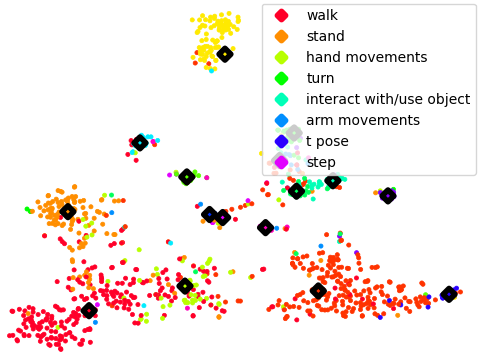}}
\end{minipage}
\hspace*{1cm}
\begin{minipage}[t]{0.2\linewidth}
\vspace{0pt}
\centering
\fbox{\includegraphics[height=1\linewidth]{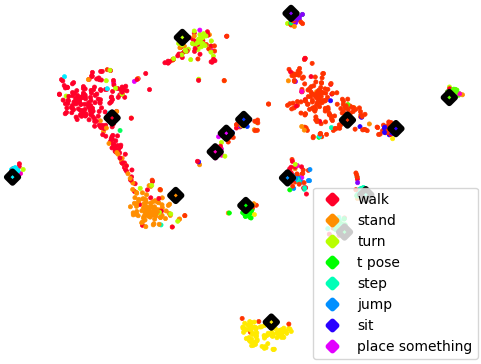}}  
\end{minipage}
\hspace*{1cm}
\begin{minipage}[t]{0.4\linewidth}
\vspace{0pt}
\centering
\resizebox{1\linewidth}{!}{
\begin{tabular}{lccc}
            \toprule
            Method & Text features & \multicolumn{2}{c}{NTURGB+D-60}\\
            \midrule
            & & top1 & top5\\
            \midrule
            & CLIP & 19.34 & 40.13 \\
            Ours & LLM + CLIP & 26.17 & 53.88\\
            \bottomrule
            \end{tabular}
            }
\end{minipage}
\caption{\textbf{Motion and text feature embeddings analysis.} Left: TSNE with features out of CLIP with the single label as text. Right: Embeddings with the augmented text from the LLM. The colored dots represent the motion features, and the diamonds the text features. We can observe that CLIP features become closer to the motion features when the text chosen as input is augmented with the LLM prompt. This is also observed in the ablation study (right), where we can notice a clear improvement with our orientation-aware LLM prompts for cross-domain scenarios.}
\label{fig:tsne}\vspace*{-0.5cm}
\end{figure}

\subsection{Ablation and Sensitivity Analysis} 
We perform different ablation studies to evaluate the multiview components in \cref{tab:classification}. The multiview projection stage during the training alone provides a large improvement in the classification results. 
The projection stage adds a top1 gain of $+4.61\%$ with 2s-AGCN encoder. The selection of this encoder backbone for comparison was also guided due to available classification results on BABEL. We can observe overall better results with ProtoGCN as encoder. Again, the orientation-aware components of our network allows a gain in both single view and multiple view inference settings, where top1 and top5 results obtain an overall augmentation of $+6.27\%$ and $+9.22\% $ respectively.
One limitation of the proposed pipeline is the requirement to estimate body orientation at inference time, which incurs additional computational overhead. Future work could address this limitation by integrating body orientation estimation directly into the pose estimation process, thereby reducing inference complexity.
The influence of the LLM is highlighted in the ablation~\cref{fig:tsne} where textual described movement increases the capability to match actions in different domains.
This effect can be also noticed in the qualitative analysis of the features shown in \cref{fig:tsne}, which demonstrates how the designed language components bring motion features closer to the text ones.

\vspace{-0.3cm}

\section{Conclusion}
{This paper presents a new human action recognition technique leveraging motion viewpoint and textual descriptions of actions in the learning of feature representations. The core interests of this approach are twofold: i) we propose an orientation-aware motion encoding network that maintains robustness under varying camera setups and human body orientations, addressing a significant source of domain shift. ii) We leverage existing action labels with text descriptions and a language module to align the motion features with different possible types of actions across domains. Several experiments and evaluations on three widely adopted action recognition benchmarks (NTU, BABEL and NW-UCLA) and two \textit{in the wild} applications (RHM-HAR and MCAD) indicate two key findings: first, significant recognition improvements can be achieved within the same domain when leveraging action viewpoint cues; secondly, cross-domain evaluations demonstrate that the proposed approach generalizes effectively to unseen motion and action classes, supporting zero-shot action recognition and highlighting its potential for real-world deployment in diverse scenarios as closed-circuit television (CCTV) footage, surveillance or human-robot interaction.}

\noindent\textbf{Acknowledgements.} This work was funded by TEB Group -- Prynel SAS, and by grants from projects ANER MOVIS from ``Conseil Regional de Bourgogne-Franche-Comte'' and ANR MANYVIS (ANR-23-CE23-0003-01), to whom we are grateful.

{\small
\bibliographystyle{splncs04}
\bibliography{biblio.bib}
}

\vfill\pagebreak

\end{document}